%% file: main_for_arxiv.tex
\pdfoutput=1

\documentclass[11pt]{article}

\usepackage[final]{acl}

\usepackage{times}
\usepackage{latexsym}

\usepackage[T1]{fontenc}

\usepackage[utf8]{inputenc}

\usepackage{microtype}

\usepackage{inconsolata}

\usepackage{graphicx}

\usepackage{comment}
\usepackage{booktabs}
\usepackage{multirow}
\usepackage{makecell}
\usepackage{xurl}
\usepackage[most]{tcolorbox}
\usepackage[table, x11names]{xcolor}
\usepackage{tabularx}
\newcolumntype{C}{>{\centering\arraybackslash}X}
\usepackage[normalem]{ulem}

\usepackage[whole]{bxcjkjatype}
\usepackage{threeparttable}

\newcommand{\return}{\mbox{\textcolor{gray}{\textbackslash n}}}
\newcommand{\pspace}{\textcolor{gray}{␣}}
\newcommand{\ptemplate}[1]{\textcolor{blue}{\texttt{\string{#1\string}}}}

\title{Is Micro Domain-Adaptive Pre-Training Effective for Real-World Operations? Multi-Step Evaluation Reveals Potential and Bottlenecks}

\author{Masaya Tsunokake \quad Yuta Koreeda \quad Terufumi Morishita \\ 
\textbf{Koichi Nagatsuka \quad Hikaru Tomonari \quad Yasuhiro Sogawa}\\
        Research \& Development Group, Hitachi, Ltd\\ Tokyo, Japan\\
 \small{
    \textbf{Correspondence:} \href{masaya.tsunokake.qu@hitachi.com}{masaya.tsunokake.qu@hitachi.com}
 }        
}

\begin{document}
\maketitle
\begin{abstract}
When applying LLMs to real-world enterprise operations, LLMs need to handle proprietary knowledge in small domains of specific operations (\textbf{micro domains}).
A previous study \cite{mdapt_xue} shows micro domain-adaptive pre-training (\textbf{mDAPT}) with fewer documents is effective, similarly to DAPT in larger domains.
However, it evaluates mDAPT only on multiple-choice questions; thus, its effectiveness for generative tasks in real-world operations remains unknown.
We aim to reveal the potential and bottlenecks of mDAPT for generative tasks.
To this end, we disentangle the answering process into three subtasks and evaluate the performance of each subtask: (1) \textbf{eliciting} facts relevant to questions from an LLM's own knowledge, (2) \textbf{reasoning} over the facts to obtain conclusions, and (3) \textbf{composing} long-form answers based on the conclusions.
We verified mDAPT on proprietary IT product knowledge for real-world questions in IT technical support operations. 
As a result, mDAPT resolved the elicitation task that the base model struggled with but did not resolve other subtasks.
This clarifies mDAPT's effectiveness in the knowledge aspect and its bottlenecks in other aspects.
Further analysis empirically shows that resolving the elicitation and reasoning tasks ensures sufficient performance (over 90\%), emphasizing the need to enhance reasoning capability.
\end{abstract}

\section{Introduction}
\begin{figure*}[tb]
    \centering
    \includegraphics[keepaspectratio, width=1.0\linewidth]{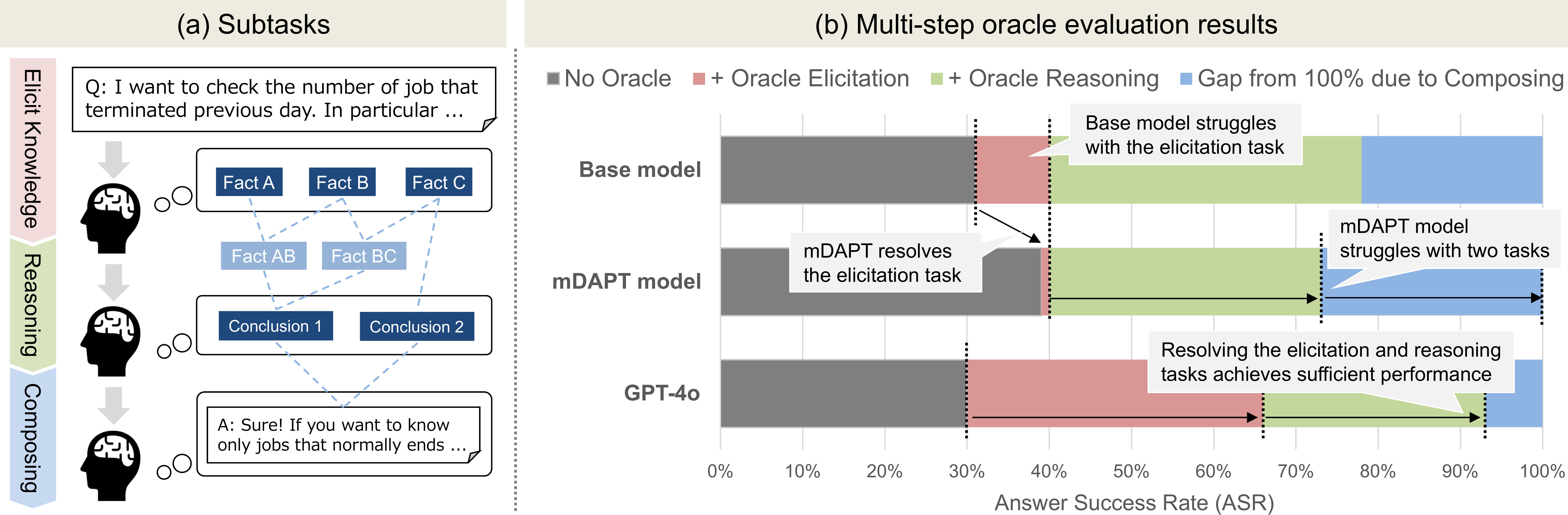}
    \caption{(a) Subtasks in an answering process (b) Results by our evaluation framework. To identify bottleneck tasks, our framework observes performance changes after inserting ideal results of each task (\textbf{oracle result}) into prompts. Although the base model struggles with the elicitation task, mDAPT resolves this difficulty, showing mDAPT's effectiveness. However, the mDAPT model still struggles with the reasoning and composing tasks.}\label{fig:intro_figure}
\end{figure*}
Driven by the remarkable advances of large language models (LLMs)~\cite{NEURIPS2020_1457c0d6,achiam2023gpt4}, many companies are increasingly utilizing LLMs for their internal operations.
When applying LLMs to real-world operations, LLMs need to handle proprietary knowledge in each company or operation~\cite{ling2023domain,zhao2024_enterpriseLLM_beyondRAG}. 
However, LLMs cannot generate content grounded in knowledge outside their training data.

Domain-adaptive pre-training (DAPT)~\cite{gururangan-etal-2020-dont} is one approach for enabling LLMs to handle unseen knowledge. 
Previous studies show DAPT improves LLMs on many tasks in medical~\cite{singhal2023expertlevel}, financial~\cite{wu2023bloomberggpt}, legal~\cite{colombo2024saullm7bpioneeringlargelanguage}, and code~\cite{gunasekar2023textbooksneed} domains.
Meanwhile, real-world operations often demand knowledge within \textbf{small} and \textbf{proprietary} domains of specific operations (\textbf{micro domains}), and micro domains have far fewer documents than larger domains.

\citet{mdapt_xue} investigated the effectiveness of DAPT in a micro domain (\textbf{mDAPT}: micro Domain-Adaptive Pre-Training).
However, their evaluation was limited to multiple-choice questions (MCQs).
Compared to MCQs where LLMs can select from limited choices, complicated real-world questions require LLMs not to select but generate long-form answers from scratch by utilizing their trained knowledge.
To advance enterprise use of LLMs, it is important to reveal whether mDAPT is effective for generative tasks in real-world operations, and if not, what bottlenecks there are.

This paper aims to reveal the potential and bottlenecks of mDAPT for generative tasks.
To clarify what aspects of generative tasks are difficult for mDAPT models, we disentangle the answering process into three subtasks: (1) \textbf{eliciting} facts relevant to questions from an LLM's own knowledge, (2) \textbf{reasoning} over the facts to obtain conclusions, and (3) \textbf{composing} long-form answers based on the conclusions (Figure~\ref{fig:intro_figure}(a)).
We identify bottleneck subtasks by observing LLM's overall performance changes after inserting an ideal result of each subtask (\textbf{oracle result}) into prompts.
Inserting a subtask's oracle result enables LLMs to solve subsequent subtasks based on it.
Thus, if inserting an oracle result of a previous subtask improves overall performance, it indicates that the LLM was not able to adequately solve the subtask by itself, and the subtask is a bottleneck.

We trained mDAPT models from Qwen2.5-72B-Instruct~\cite{qwen25technicalreport} with proprietary IT product knowledge and evaluated them on real-world questions in IT technical support operations.
We found that mDAPT resolves the elicitation task that the base model struggles with (Figure~\ref{fig:intro_figure}(b)), showing mDAPT's effectiveness in acquiring and eliciting knowledge.
However, the mDAPT model exhibits insufficient performance (39\%). 
The improvements by inserting oracle results show that the mDAPT model struggles with the reasoning and composing tasks, revealing mDAPT's bottlenecks.

For comparison, Figure~\ref{fig:intro_figure}(b) also shows the performance of GPT-4o~\cite{openai2024gpt4ocard}.
This indicates that sufficient performance (over 90\%) could be achieved by resolving the elicitation and reasoning tasks if a stronger proprietary model were the base model.
Given mDAPT resolves the elicitation task, our results empirically highlight that enhancing reasoning capability to resolve the reasoning task is a high-priority and promising approach for realizing a sufficiently usable model for real-world operations.

\section{Background\label{sec:background}}
\subsection{Micro Domain} \label{subsec:define_micro_domain}
A previous study focused on micro domain knowledge of an IT product for evaluating mDAPT~\cite{mdapt_xue}. 
Since many companies introduce IT products for their operations, including proprietary ones, question answering for IT products is a representative use case in real-world operations.
Therefore, we adopt this use case to study mDAPT.

\begin{figure}[t]
    \centering
    \fontsize{7.6pt}{7.2pt}\selectfont
    \begin{tcolorbox}[before upper=\raggedright]
\textbf{3.1 Hierarchical structure of the job network}\\
\quad In JP1/AJS3, the elements in an application to be automated are known as \textit{units}.
\dashuline{Each of the individual processes involved in an application is defined as a unit called a \textit{job}. A job is the smallest unit}. You then arrange the defined jobs in order of execution, creating a network of jobs grouped together to form a unified application. This collection of jobs is called a \textit{jobnet}.
\tcbline
\textbf{2.2.5 Using wait conditions to control the order of unit execution}\\
\quad You can use a \dashuline{\textit{wait condition} to control the execution sequence of units} in different jobnets. A unit assigned a wait condition does not start to execute until a specified unit ... 
\end{tcolorbox}
\caption{Facts written in JP1 manuals~\cite{JP1_manuals}}\label{fig:fact_knowledge_sample}
\end{figure}

\begin{figure*}[tb]
    \centering
    \includegraphics[keepaspectratio, width=0.98\linewidth]{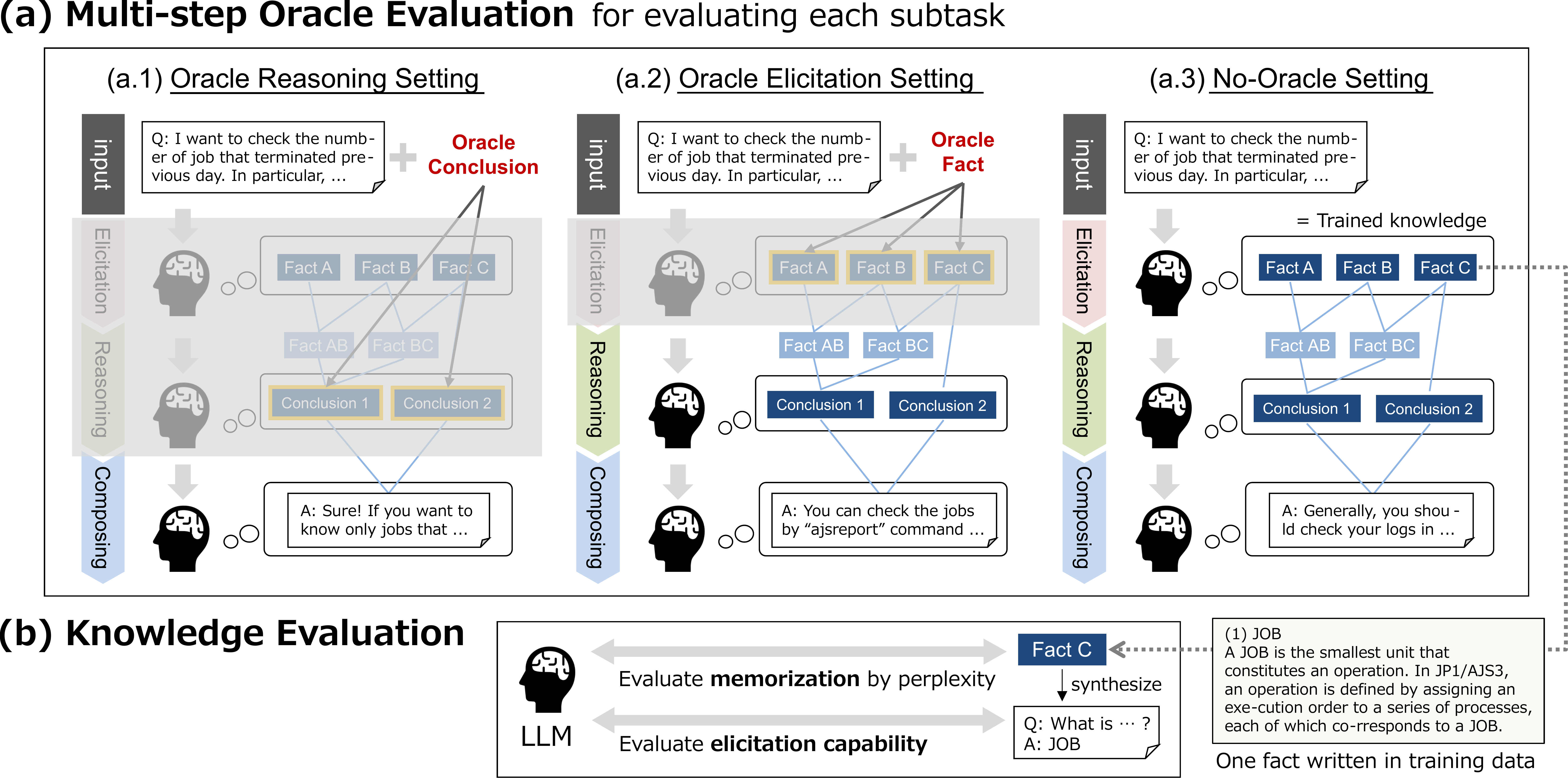}
    \caption{Our evaluation framework consists of multi-step oracle evaluation and knowledge evaluation. In multi-step oracle evaluation, we observe LLM's overall performance changes after inserting oracle results for subtasks. If the performance is improved, it means that an LLM could not solve the corresponding task by itself. In knowledge evaluation, we evaluate LLM's memorization and elicitation capability by using trained texts relevant to questions.}\label{fig:overview_analysis_framwork}
\end{figure*}
We use JP1~\cite{about_JP1}, a typical IT product that constitutes proprietary domain knowledge, for mDAPT as in the previous study~\cite{mdapt_xue}.
JP1 consists of several software (e.g., job scheduler, database) with proprietary terminology and specification sets.
Figure~\ref{fig:fact_knowledge_sample} shows JP1's proprietary terms.
According to the upper example, the term ``job'' has a definition that differs from its general meaning.
The lower example describes ``waiting condition,'' which is a proprietary specification.
Reasoning over underlined facts yields the new conclusion that ``\dashuline{a wait condition can control the execution sequence of \textbf{jobs}}.''
As like this, various facts mutually interact in this JP1 domain.

\subsection{Micro Domain-Adaptive Pre-Training}
Our mDAPT consists of continual pre-training (CPT) and supervised fine-tuning (SFT), which is a standard DAPT procedure.
CPT is performed by a next token prediction task on raw corpora.
SFT is performed by the same task, using only a subset of tokens for loss computation. We use QA pairs for SFT; thus, the loss is calculated only on answer tokens. We synthesize the QA pairs from raw corpora used for CPT to enhance LLMs' capability to utilize knowledge~\cite{cheng-etal-2024-instruction,jiang-etal-2024-instruction,cheng2024adapting,ziegler2024craftdatasettaskspecificsynthetic}.

\section{Evaluation Framework} \label{sec:eval_framework}
\subsection{Overview} \label{subsec:overview_eval_framework}
\begin{table*}[t]
    \centering
    \fontsize{7.7pt}{9pt}\selectfont

    \begin{threeparttable}
    
    \setlength{\tabcolsep}{4pt}
    {\renewcommand\arraystretch{1.4}
    \begin{tabular}{lll}
    \specialrule{\heavyrulewidth}{\aboverulesep}{0pt}
        Usage & Data name & Example \\ \hline
        \multirow{4}{*}{\makecell[tl]{Multi-step\\Oracle\\Evaluation}} & \makecell[tl]{Condition for\\LLM-as-a-judge} & \makecell[tl]{The given answer describes a solution that uses the ``ajsreport'' command to output the previous day's perform-\\ance report and aggregates ``JOB\_EXEC\_END\_N\_NUM,'' ``NEST\_EXEC\_END\_N\_NUM,'' and ``RJ\_EXEC\_\\END\_N\_NUM.''} \\ \cline{2-3}
        & \makecell[tl]{Oracle conclusion} & \makecell[tl]{By executing the ``ajsreport'' command to output the previous day's performance report and checking ``JOB\_E\\XEC\_END\_N\_NUM,'' ``NEST\_EXEC\_END\_N\_NUM,'' and ``RJ\_EXEC\_END\_N\_NUM'' items, you can con-\\firm the number of jobs and jobnets that terminated normally previous day with a single command.} \\ \cline{2-3}
        & \makecell[tl]{Oracle fact 1} & \makecell[tl]{\textit{ajsreport}\quad The ajsreport command outputs JP1/AJS3 performance reports.} \\ \cline{2-3}
        & \makecell[tl]{Oracle fact 2} & \makecell[tl]{NEST\_EXEC\_END\_N\_NUM: Number of nested jobnets or nested remote jobnets that terminated normally.\\This value is incremented according to the number of jobnets that are placed in the Ended normally status.} \\ \hline
        \multirow{2}{*}{\makecell[tl]{Knowledge\\Evaluation}} & \makecell[tl]{Closed-book QA\\synthesized from\\oracle fact 2} & \makecell[tl]{Q: Which parameter indicates the number of nested jobnets or nested remote jobnets that have terminated nor-\\mally, where the value increases according to the number of jobnets in the ``Ended normally'' status?\\A: NEST\_EXEC\_END\_N\_NUM} \\
    \specialrule{\heavyrulewidth}{0pt}{\belowrulesep}
    \end{tabular}
    }
    \end{threeparttable}
    \caption{Example data used in our evaluation framework. We show examples made by authors that replicate the characteristics of our real data due to confidentiality issues. The condition for LLM-as-a-judge specifies information that LLMs' answers should cover. The oracle conclusion is information that directly leads to the correct answer. The oracle facts are extracted training data~\cite{JP1_ajs_manuals} and support the oracle conclusion.}  \label{tab:Exampe_eval_data}
\end{table*}

\begin{table*}[t]
    \centering
    \scriptsize
    
    \setlength{\tabcolsep}{7pt}
    \begin{tabular}{llllrrlrl}\toprule
        \multicolumn{2}{c}{Name} & \multicolumn{1}{c}{Usage} & \multicolumn{1}{c}{Description} & \multicolumn{1}{c}{\# Documents} & \multicolumn{2}{c}{Size (MB)} & \multicolumn{2}{c}{\# Tokens (M)} \\\midrule
        \multicolumn{2}{l}{JP1 Documents}    & CPT & \makecell[tl]{JP1 manuals~\cite{JP1_manuals}, release notes, and other references} & 193 & 72.1 & & 18.3 & \\
        \multicolumn{2}{l}{llm-jp-corpus-v2}    & CPT & \makecell[tl]{Subset of llm-jp-corpus-v2, which contains various Japanese texts} & - & 651.4 &  & - & \\
        \multicolumn{2}{l}{JP1-QA}            & SFT & JP1/AJS-related QA pairs synthesized from JP1 manuals & 12,305 & 42.9 & (9.5) & 10.6 & (2.1) \\
        \bottomrule
    \end{tabular}

    \caption{Training data of mDAPT. The JP1 documents are obtained by converting original PDF files to texts. For JP1-QA, the statistics of the answer part, actually used for the SFT loss computation, are shown within parentheses.} \label{tab:data_statistics}
\end{table*}

LLMs generally need to generate answers using their knowledge.
Thus, LLMs should be capable of \textbf{eliciting} knowledge.
However, LLMs may receive questions asking about unseen facts not directly written in trained knowledge.
In such scenarios, LLMs must \textbf{reason} over known facts to obtain new conclusions needed for answers, a well-known nature in multi-hop question answering~\cite{hotpotqa,musique_data}.
In enterprise use, LLMs need to handle multi-hop questions grounded in proprietary knowledge~\cite{zhao2024_enterpriseLLM_beyondRAG}.
Therefore, we claim that mDAPT models need to address three subtasks in an answering process: (1) \textbf{eliciting} relevant facts from their trained knowledge, (2) \textbf{reasoning} over the facts to obtain conclusions, and (3) \textbf{composing} long-form answers based on the conclusions, as shown in Figure~\ref{fig:overview_analysis_framwork}(a.3).

We aim to clarify whether each subtask is a bottleneck in real-world questions.
To this end, we introduce a simple but explainable method that observes LLM's overall performance changes after inserting an ideal result of each subtask (\textbf{oracle result}) into prompts.
For example, inserting an oracle result of the reasoning task (\textbf{oracle conclusions}) allows us to observe an LLM's overall performance when the reasoning task is perfectly solved.
If inserting oracle conclusions improves the overall performance, it indicates that the LLM was not able to adequately solve the reasoning task by itself, and the reasoning task is a bottleneck.
We call this evaluation \textbf{Multi-step Oracle Evaluation}.

The elicitation task assumes that LLMs have memorized the knowledge and can access them on demand.
To verify whether these requirements are bottlenecks, we additionally evaluate LLMs' memorization and how well LLMs can access knowledge (\textbf{elicitation capability}).
We call this evaluation \textbf{Knowledge Evaluation}.

\subsection{Multi-step Oracle Evaluation} \label{subsec:parallel_oracle_eval}
\subsubsection{Multiple Oracle Settings}
We define the following three prompt settings to realize the multi-step oracle evaluation (Figure~\ref{fig:overview_analysis_framwork}(a)).

\paragraph{Oracle Reasoning Setting}
This inserts \textbf{oracle conclusions}, information directly leading to correct answers, into prompts. This allows LLMs to compose answers based on ideal results of the reasoning task (Figure~\ref{fig:overview_analysis_framwork}(a.1)).
We can determine whether the composing task is a bottleneck by comparing this performance with the upper bound.
Oracle conclusions include domain-specific facts mentioned in correct answers and guidance information that specifies how LLMs should construct answers. 
Table~\ref{tab:Exampe_eval_data} shows an example of an oracle conclusion.

\paragraph{Oracle Elicitation Setting}
This inserts oracle results of the elicitation task, relevant texts from training data (\textbf{oracle facts}), into prompts.
Thus, LLMs can address only the reasoning and composing tasks based on oracle facts (Figure~\ref{fig:overview_analysis_framwork}(a.2)).
We can determine whether the reasoning task is a bottleneck by comparing performances of this and the oracle reasoning settings. Table~\ref{tab:Exampe_eval_data} shows examples of oracle facts. 

\paragraph{No-Oracle Setting}
This does not insert oracle results into prompts; thus, LLMs must address all tasks.
We can determine whether the elicitation task is a bottleneck by comparing performances of this and the oracle elicitation settings.

\subsubsection{LLM-as-a-judge for Each Setting}
To efficiently evaluate each setting, we introduce an LLM-as-a-judge~\cite{first_llm_as_a_judge} method that determines whether each generated answer covers all required information.
Given multiple \textbf{checklists}\footnote{Each question may be associated with more than one checklist as there can be multiple approaches to answering.} each of which contains \textbf{conditions} for one question to be deemed correct, an evaluator LLM judges whether a generated answer satisfies each condition.
If the generated answer satisfies all conditions in any checklist, the answer is regarded as correct.
Table~\ref{tab:Exampe_eval_data} shows an example of a condition prompted into an evaluator LLM.

\subsection{Knowledge Evaluation} \label{subsec:knowl_eval}
Based on previous studies~\cite{jiang-etal-2024-instruction,chang_2025_how_do_LLM}, we evaluate \textbf{memorization} by computing perplexity of oracle facts and evaluate \textbf{elicitation capability} by computing accuracy on closed-book QAs synthesized from the oracle facts.
Figure~\ref{fig:overview_analysis_framwork}(b) shows the procedure.
A QA pair is synthesized by paraphrasing a given oracle fact so that one noun phrase in the fact serves as the answer.
Table~\ref{tab:Exampe_eval_data} shows an example of a QA pair.

\begin{figure*}[t]
 \centering
 \includegraphics[keepaspectratio, width=1.0\linewidth]{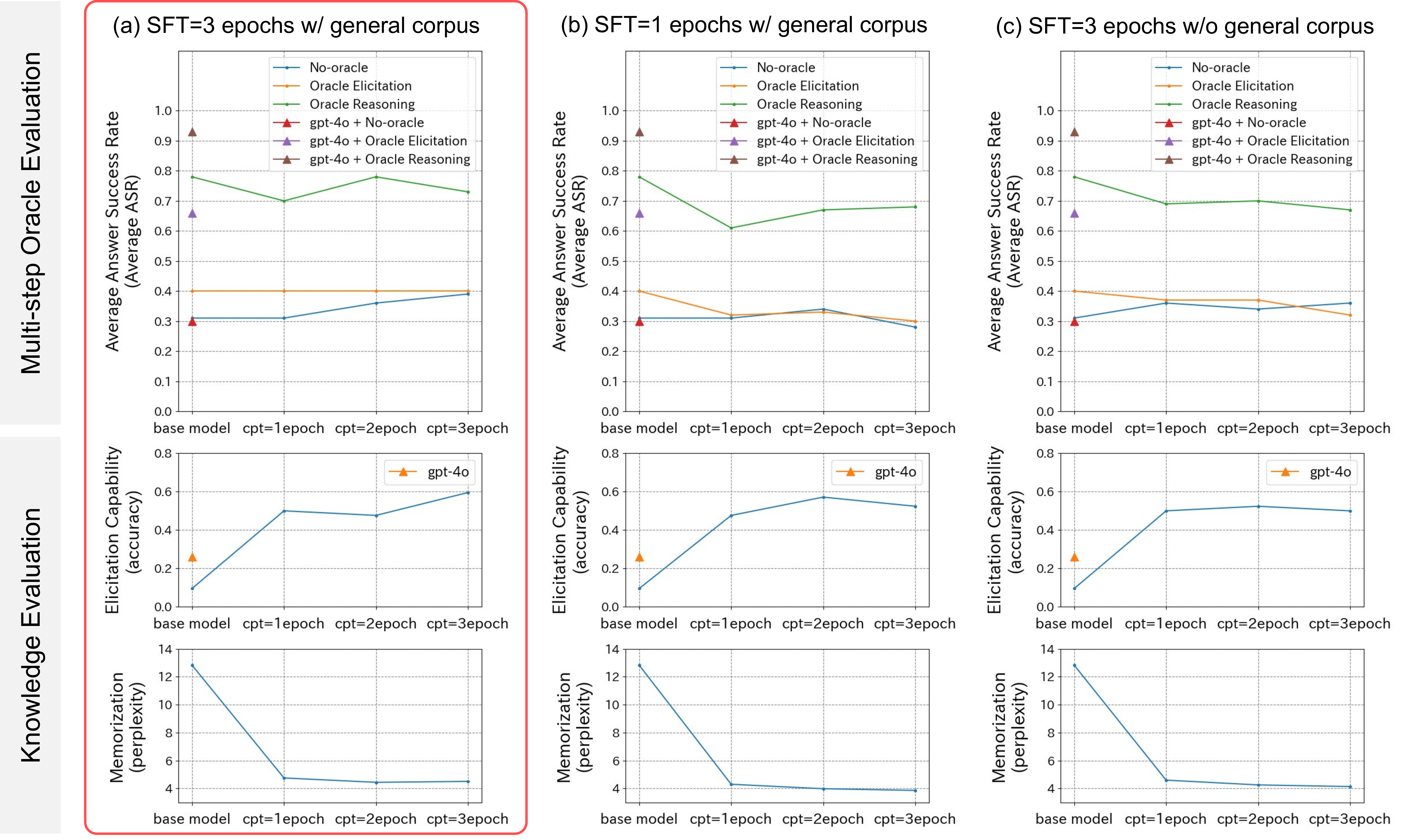}
 \caption{Evaluation results. The main results on the far-left show that memorization and elicitation capabilities improve as CPT epochs increase. At the third epoch, the ASR of the mDAPT model on the no-oracle setting reach the base model's ASR on the oracle elicitation setting, indicating that mDAPT resolves the elicitation task.}\label{fig:main_result}
\end{figure*}

\section{Experiment} \label{sec:exp}
\subsection{Experimental Setting} \label{subsec:exp_setting}

\subsubsection{Training Data} \label{subsubsec:train_data}
Table~\ref{tab:data_statistics} shows our training data.
As for CPT, We used (i) Japanese JP1 documents and (ii) a randomly sampled subset of llm-jp-corpus-v2~\cite{llm_jp_corpus_v2}, a Japanese corpus that covers diverse sources. 
Our SFT data is QAs synthesized from sampled JP1 manuals by \textit{gpt-4-1106-preview}. 
Figure \ref{fig:synthesis_prompt_jp1QA} shows the prompt for synthesis.

\subsubsection{Evaluation Setting} \label{subsubsec:eval_setting}

\paragraph{Real-world QA Data} \label{para:real_qa_data}
We evaluate mDAPT on ten real-world technical support questions.
They contain JP1 customers' questions which were too difficult for service desk personnel to handle and have to be escalated to JP1 experts.

\paragraph{Multi-step Oracle Evaluation Data}
The checklists for LLM-as-a-judge were created through interviews with a JP1 expert.
Oracle conclusions were created by writing JP1-related facts based on the checklists.
Oracle facts were created by extracting texts that support the oracle conclusions from JP1 manuals. We show more details in Appendix~\ref{sec_appendix: detail_evaluation}.

\paragraph{LLM-as-a-judge Setting}
For each prompt setting, we generate multiple answers with ten different random seeds at a temperature of 0.7.
After that, we compute the rate of answers judged as correct by our LLM-as-a-judge (\textbf{Answer Success Rate}; \textbf{ASR}).
We used Qwen2.5-72B-Instruct~\cite{qwen25technicalreport} as the evaluator.
In a preliminary evaluation, there were only three cases where our LLM-as-a-judge contradicted a JP1 expert's evaluation among 100 generated answers. 

\paragraph{Knowledge Evaluation}
We synthesized QA pairs with GPT-4o~\cite{openai2024gpt4ocard} from the oracle facts.
To evaluate the elicitation capability, we generate answers by greedy decoding and compute accuracy based on exact matches.

\subsubsection{Models}
We performed mDAPT on Qwen2.5-72B-Instruct, which has high Japanese capability.
For comparison, we evaluated GPT-4o, which is expected to have higher composing and reasoning capabilities.

\subsubsection{mDAPT setting}
We implemented CPT and SFT with Hugging Face\footnote{https://pypi.org/project/transformers/} and TRL\footnote{https://pypi.org/project/trl/} libraries.
On both CPT and SFT, learning rates were 1.0e-5, and global batch sizes were 120.
Table~\ref{tab:train_prams} presents other training hyper-parameters.
Optimizer was AdamW~\cite{Adam}.
We used DeepSpeed ZeRO-3~\cite{deepspeed_zero} and H100 GPU $\times$ 24 for training.

\subsection{Main Result} \label{subsec:main_result}
Figure~\ref{fig:main_result}(a) shows the main result.
As the number of CPT epochs increases, both memorization and elicitation capabilities improve.
This shows that mDAPT encourages the LLM to memorize necessary knowledge and to access it when it is directly asked.
In the multi-step oracle evaluation, ASR on the no-oracle setting improves as the number of CPT epochs increases.
At the third epoch when the elicitation capability peaked, the no-oracle setting's ASR matched that of the base model on the oracle elicitation setting.
This indicates that mDAPT resolves the base model's bottleneck caused by the elicitation task.

The mDAPT model's ASRs significantly improved on the oracle reasoning setting compared to the no-oracle setting.
However, there is still a gap to 100\%.
This means that the mDAPT model struggles with both the reasoning and composing tasks.
Thus, mDAPT cannot contribute to the reasoning and composing tasks although it is effective in the elicitation task. 
Such bottlenecks limit mDAPT's effectiveness in real-world operations.

\subsection{Discussion}
\paragraph{A Key condition for achieving sufficient performance is to resolve the elicitation and reasoning tasks:} \label{paragprah:future_approach}
As shown in Figure~\ref{fig:main_result}, GPT-4o's ASR improves with the oracle elicitation and reasoning, revealing bottlenecks in the elicitation and reasoning tasks.
Moreover, the ASR on the oracle reasoning setting achieves sufficient performance (over 90\%).
This empirically reveals that resolving both tasks is a necessary and sufficient condition for achieving a usable LLM in real-world operations if the base model is a stronger model.
Given that mDAPT can resolve the elicitation task, exploring how to enhance reasoning capability to resolve the reasoning task is a promising research direction.

\paragraph{SFT and using general-domain corpora are essential for preserving composing capability:} \label{paragprah:other_mdapt_settings}
To investigate how SFT affects performance, we changed SFT epochs from 3 (Figure~\ref{fig:main_result}(a)) to 1 (Figure~\ref{fig:main_result}(b)).
The perplexity for oracle facts in Figure~\ref{fig:main_result}(b) is better than that in Figure~\ref{fig:main_result}(a) across all CPT epochs.
Meanwhile, the highest elicitation capability in Figure~\ref{fig:main_result}(b) is slightly worse than that in Figure~\ref{fig:main_result}(a).
This suggests that SFT encourages models to generalize from memorizing knowledge to leveraging it, consistent with a previous study~\cite{jiang-etal-2024-instruction}.
The ASRs for the oracle reasoning results in Figure~\ref{fig:main_result}(b) are lower than those in Figure~\ref{fig:main_result}(a) across all CPT epochs.
This indicates that SFT is beneficial for preserving the base model's composing capability required for the composing task.

Figure~\ref{fig:main_result}(c) shows evaluation results obtained without using llm-jp-corpus-v2 during training.
The perplexity for oracle facts in Figure~\ref{fig:main_result}(c) is slightly better than that in Figure~\ref{fig:main_result}(a) across 2nd or 3rd CPT epochs.
This indicates that incorporating general-domain corpora into the training data slightly hinders memorization of JP1 facts.
On the other hand, the ASRs in Figure~\ref{fig:main_result}(c) on the oracle reasoning setting are lower than those in Figure~\ref{fig:main_result}(a) over all CPT epochs.
Thus, using general-domain mitigates catastrophic forgetting of composing capability, which ensures performance over an entire answering process, and this positive effect surpasses the negative effect on memorization degradation.

\paragraph{Human Evaluation} \label{paragprah:human_eval}
\begin{table}[t]
    \centering
    \small
    \begin{threeparttable}
    
    \setlength{\tabcolsep}{2pt}
    \begin{tabularx}{\linewidth}{ll*{4}{C}}
    \toprule
    \multicolumn{2}{c}{Model} & \multicolumn{4}{c}{\# QAs with each score} \\
    \cmidrule(lr){3-6}
     & & S1\tnote{1} & S2\tnote{2} & S3\tnote{3} & S4\tnote{4} \\
     \midrule
    \multicolumn{6}{@{}l}{\textit{(a) Evaluation on 20 QAs}} \\
     \quad\quad & Qwen2.5-72B-Instruct & 13 & 6 & 1 & 0 \\
     & ~~~+ RAG & 13 & 5 & 2 & 0 \\
     & GPT-4o & 12 & 6 & 2 & 0 \\
     & mDAPT model & 10 & 9 & 1 & 0 \\
    \midrule
    \multicolumn{6}{@{}l}{\textit{(b) Evaluation on 10 QAs used in Figure~\ref{fig:main_result}'s experiment}}\\
     & mDAPT model & 4 & 6 & 0 & 0 \\
     & ~+ Oracle reasoning & 1 & 5 & 1 & 3 \\
    \bottomrule
    \end{tabularx}

    \begin{flushleft}
    \begin{tablenotes}
      \fontsize{8.5pt}{8pt}\selectfont
      \item[1] A given answer is \textbf{not useful}.
      \item[2] A given answer contains misinformation but is \textbf{useful}.
      \item[3] A given answer is \textbf{very useful} but needs some modifications.
      \item[4] A given answer \textbf{can be adopted} for the actual answer as is.
    \end{tablenotes}
    \end{flushleft}

    \end{threeparttable}

\caption{Human Evaluation by an expert} \label{tab:human_eval}
\end{table}
We also asked JP1 experts to evaluate LLMs.
Since human evaluation is time-consuming, we selected the mDAPT model with 3 CPT epochs (Figure~\ref{fig:main_result}(a)), which achieved the highest ASR on the no-oracle setting.
We used twenty technical support questions, including the ones described in Section~\ref{para:real_qa_data}.
We asked experts to categorize the LLMs' answers into four scores based on the usefulness for writing actual answers.

Table~\ref{tab:human_eval}(a) shows the result.
The mDAPT model outperformed RAG~\cite{NEURIPS2020_6b493230,gao2024retrievalaugmentedgenerationlargelanguage} and GPT-4o on the number of useful answers (scores of S2 or higher).
This indicates that mDAPT is useful in real-world operations to some extent.
We further focused on the ten QAs used in the multi-step oracle evaluation and asked experts to evaluate the answers generated in the no-oracle and oracle reasoning settings.
As shown in Table~\ref{tab:human_eval}(b), scores in the oracle reasoning setting are distributed at a higher position than those in the no-oracle setting.
This emphasizes that the reasoning task is a bottleneck, consistent with Section~\ref{subsec:main_result}.

\section{Conclusion}
We evaluated mDAPT in real-world operations that require micro domain knowledge.
mDAPT can resolve the elicitation task, but cannot resolve other tasks, which limits mDAPT's effectiveness.
Further analysis revealed that sufficient performance can be achieved by additionally resolving the reasoning task. 
Applying mDAPT to pretrained reasoning models while somehow avoiding catastrophic forgetting is one direction to addressing this challenge, yet such methods are not sufficiently explored.

\section*{Limitations}
In this paper, we clarified both the potential and the bottlenecks of mDAPT through careful multi-perspective evaluation using real-world data from professional operations.
While our evaluation framework is language-independent and can be applied to a wide range of tasks, we evaluated LLMs only on technical support QA data.
However, we believe that conducting and sharing an in-depth evaluation on proprietary real-world data has substantial value even when the results come from a single domain.
In proprietary business domains, it is often difficult to release data publicly, which makes it challenging to accumulate quantitative findings regarding the effectiveness of LLMs in real-world operations.
Therefore, we consider it important to accumulate quantitative findings even from a single domain, and our paper offers valuable and meaningful results in this respect.
By sharing both our findings and the evaluation framework, we hope to enable other organizations to conduct similar analyses in different business domains, thereby accelerating the accumulation of findings and insights about challenges related to LLM deployment across industry and academia.

The main goal of this study is to establish a framework for clarifying bottlenecks of current mDAPT, and many DAPT attempts use non-reasoning models.
Therefore, we adopted Qwen2.5-72B-Instruct, which is a non-reasoning model with top-level Japanese capability, as a base model.
However, evaluating mDAPT on more recent non-reasoning models or reasoning models would be an interesting direction for future work.
In the latter direction, we should avoid catastrophic forgetting when applying mDAPT to reasoning models. However, reasoning models are generally constructed through specialized training to enhance reasoning capability~\cite{deepseekR1,s1simpletesttimescaling2025,ProRL2025}, and mDAPT on such reasoning models keeping reasoning capability is still under-explored.

Moreover, many approaches for training reasoning models focus on developing logical thinking for mathematical problems, rather than on reasoning capability over proprietary knowledge.
Thus, whether such models' reasoning capabilities resolve our defined reasoning sub-task is not trivial.
Training methods for the latter purpose are still under development in large proprietary domains~\cite{huang2025m1unleashpotentialtesttime,wu2025medreasonelicitingfactualmedical,FinR12025}.
If well-established methods become available, it would be desirable to apply such training method instead of mDAPT to available LLMs and evaluate the trained models using our framework.

To evaluate standard mDAPT approach, we finetuned all parameters  of LLMs by CPT and SFT procedures.
This requires large machine resources, H100 GPU $\times$ 24 for both CPT and SFT in our experiments, and makes practical adoption difficult.
Thus, we further would like to verify light-weight training methods such as LoRA~\cite{lora} in future work.

\section*{Acknowledgments}
We would like to thank anonymous reviewers and Kana Ozaki for their valuable feedback. We would also like to thank Dr. Masaaki Shimizu for the maintenance and management of large computational resources. Additionally, we would also like to thank Hitachi Solutions, Ltd. for their support of this research.

\bibliography{refs}

\newpage
\appendix

\begin{table*}[t]
    \centering
    \scriptsize
    
    \setlength{\tabcolsep}{5pt}
    \begin{tabular}{llllrrlrl}\toprule
        \multicolumn{2}{c}{Name} & \multicolumn{1}{c}{Usage} & \multicolumn{1}{c}{Description} & \multicolumn{1}{c}{\# Documents} & \multicolumn{2}{c}{Size (MB)} & \multicolumn{2}{c}{\# Tokens (M)} \\\midrule
        \multicolumn{2}{l}{JP1 Manuals~\cite{JP1_manuals}}       & CPT & \makecell[tl]{Text converted from PDF manuals covering JP1 Version 13.} & 105 & 56.8 & & 14.3 & \\
        \multicolumn{2}{l}{JP1 Release Notes} & CPT & \makecell[tl]{Text converted from PDF release notes of JP1.} & 42 & 14.9 & & 3.9 & \\
        \multicolumn{2}{l}{Other JP1 Docs}    & CPT & \makecell[tl]{Other references for managing JP1} & 46 & 0.4 & & 0.1 & \\
        \multicolumn{2}{l}{llm-jp-corpus-v2}    & CPT & \makecell[tl]{Subset of llm-jp-corpus-v2, which contains various Japanese texts} & - & 651.4 &  & - & \\
        \bottomrule
    \end{tabular}

    \caption{Detailed list of training data used for CPT} \label{tab:detail_cpt_data}
\end{table*}

\section{Detailed Training Settings} \label{sec_appendix:train_data}
\input{appendix/prompt_synthesize_sftdata}
Table~\ref{tab:detail_cpt_data} provides a detailed breakdown of the training data used for CPT.
JP1 documents, our micro domain-specific corpora, consist of JP1 manuals~\cite{JP1_manuals}, JP1 release notes, and other JP1 references in Japanese.
All JP1 documents were created by converting original PDF files to texts by PDFminer\footnote{\url{https://pypi.org/project/pdfminer/20191125/}}.

As for SFT data, Figure \ref{fig:synthesis_prompt_jp1QA} shows a prompt used for synthesizing JP1-QA, our SFT data, from JP1 manuals. 
\begin{table}[t]
    \centering
    \small
        
    \setlength{\tabcolsep}{5pt}
    \begin{tabular}{ccc}
        \toprule
        \multicolumn{1}{c}{Parameter} & \multicolumn{2}{c}{Value}\\
        \cmidrule(lr){2-3}
         & \multicolumn{1}{c}{CPT} & \multicolumn{1}{c}{SFT} \\
        \midrule
        Global batch size & 120 & 120 \\
        Learning rate & 1.0e-5 & 1.0e-5 \\
        Weight decay & 0.1 & 0.1 \\
        Adam $\beta$ 1 & 0.9 & 0.9 \\
        Adam $\beta$ 2 & 0.95 & 0.95 \\
        Optimizing Scheduler & constant & - \\
        Warm up ratio & - & - \\
        Max sequence length & 2048 & 4096 \\
        \bottomrule
    \end{tabular}

    \caption{Training hyper-parameters} \label{tab:train_prams}
\end{table}

We sequentially performed CPT and SFT in mDAPT. Table~\ref{tab:train_prams} shows detailed training hyper-parameters. We used H100 GPU $\times$ 24 for both CPT and SFT. 
The mDAPT model on CPT three epochs shown in the left side of Figure~\ref{fig:main_result}, which has the highest ASR on the no-oracle setting, was trained by 80 hours.

\section{Evaluation Data} \label{sec_appendix: detail_evaluation}
\paragraph{Real-world QA Data}
QA pairs used in our experiments are from real-world technical support operations. However, personal information has been manually anonymized, and e-mail histories leading up to questions have been manually removed.

\paragraph{Multi-step Oracle Evaluation Data}
The checklists for LLM-as-a-judge were constructed through interviews with JP1 experts, who actually work for technical support operations. The average number of checklists per QA pair is 1.8. 
The average number of conditions per checklist is 2.6.

Oracle conclusions were created by writing JP1-related facts based on the checklists' conditions.
Some of them constitute ground-truth answers, thus, LLMs can generate correct answers given oracle conclusions if the LLMs have sufficient reading and generation capabilities.
In a pilot study, \textit{gpt-4-0613} generated correct answers for seven out of ten QA pairs by inserting the oracle conclusions into the prompts.
We found that the remaining three QA pairs need not only factual knowledge, but also domain-specific thinking patterns based on factual knowledge to generate correct answers.
Thus, we appended guidance information (how to make plausible explanation, where to emphasize, etc.), described in Section~\ref{subsec:parallel_oracle_eval}, to the oracle conclusions. 

\input{appendix/prompt_synthesize_factoidQA}
Oracle facts were created by searching rationale texts for oracle conclusions from JP1 manuals and extracting them.
Texts extracted from the same section were unified to one oracle fact by concatenating the texts.
After that, we appended section titles to a part of oracle facts.
There are 4.6 oracle facts per QA pair.
A JP1 expert evaluated all created oracle facts as being relevant to questions and 74\% of oracle facts as being mandatory knowledge for writing answers.

\paragraph{Knowledge Evaluation}
For evaluating memorization, we computed perplexity for each paragraph in the oracle facts and report the average values.
For evaluating elicitation capability, we synthesized closed-book QA pairs from the oracle facts by GPT-4o.
Figure~\ref{fig:synthesis_prompt_factoidQA} shows a prompt used for synthesizing. The example of ``\#\#\# Document'' in Figure~\ref{fig:synthesis_prompt_factoidQA} is cited from \url{https://itpfdoc.hitachi.co.jp/manuals/3021/30213L4210e/AJSF0060.HTM}.
After synthesizing by this prompt, we manually corrected or deleted synthesized questions that have multiple possible answers.
Consequently, 42 QAs were created, and we used them in experiments.

\section{Evaluation Details}
\input{appendix/prompt_guidance_information}
\input{appendix/prompt_llm_as_a_judge}
\input{appendix/prompt_eval_elicit_capability}
When generating answers for technical support questions, we simply asked LLMs to answer given questions in prompts.
On the oracle reasoning setting, we insert ``\#\# Background Knowledge'' and itemized oracle conclusions into the prompts before the questions. If the oracle conclusions include guidance information that specifies LLMs' reasoning direction, we additionally insert texts shown in Figure~\ref{fig:prompt_guidance_info} after the questions. On the oracle elicitation setting, we insert ``\#\# Background Knowledge'' and oracle facts concatenated with ``\#\#\#\# Knowledge {\textit{i}},'' where \textit{i} is an index of each oracle fact. We actually used Japanese prompts in the experiments, but we also show their English translation for explanation purposes in this paper.

Our LLM-as-a-judge focuses on the correctness of generated answers. Ideally, a correct answer should cover all required information, and it does not contain misinformation or irrelevant content. Nevertheless, it is difficult to determine the latter as there is an immense variety of patterns. Accordingly, we focus on the former; evaluating whether generated answers satisfy essential conditions for correctness.
Figure \ref{fig:prompt_llm_as_a_judge} shows a prompt used for LLM-as-a-judge. If an evaluator LLM outputs ``Yes'' or ``yes'' for this prompt, the generated answer is judged to satisfy a given condition. This judgment process is applied to all conditions in all prepared checklists.

In the knowledge evaluation, we evaluated LLM's elicitation capability by a prompt shown in Figure~\ref{fig:prompt_eval_elicit_capability}.
Generated answers are first normalized.
Specifically, we split each generated text by newline character and normalize the first part of split texts by removing punctuation from it.
After normalization, we computed accuracy based on whether normalized answers exactly match the ground-truth answers.

\end{document}

%% file: appendix/prompt_synthesize_sftdata.tex
\begin{figure}[t]
    \centering
    \fontsize{7.5pt}{7.5pt}\selectfont

    \begin{tcolorbox}
以下の文書はソフトウェア製品のマニュアルです。\return\\
次の手順でこの文書に関する質問、回答、回答の根拠を日本語で作成してください。\return\\
\return\\
1.\pspace この製品を使うユーザーになりきって、Question:という文字列の後に、与えられた文書に基づいて、実際に起きたトラブルや聞きたいこと、分からないことなどを質問として作成して下さい。その際、トラブルに加えて、やりたいことや、状況説明、トラブルシューティングに役立つ情報(例えば、実行環境、バージョン、実行コマンド、エラー詳細)なども記載するといいかもしれません。\return\\
2.\pspace 次に、この製品を扱う会社の一流のコンタクトセンターの職員になりきって、Answer:という文字列の後に、文書に基づいて初心者のユーザーの質問に丁寧に答えて下さい。\return\\
3.\pspace 最後に、Citation:という文字列の後に、回答の根拠となった文書内の記述をそのままコピー\&ペーストして書いてください(変更は加えないで抜き出してください)。\return\\
\return\\
注意点として、質問や回答を複数個つくったりしないでください。また、Question:, Answer:, Citation:以外のフォーマットを使わないでください。 \return\\
\return\\
では、質問を以下の文書の情報を基に日本語で作成してください。\return\\
文書:\ptemplate{chunk}
\tcblower
\textcolor{gray}{(A prompt translated into English)}\\
The following document is a manual for a software product.\return\\
Please create a question about this document, the answer, and the rational of the answer in Japanese according to the steps below.\return\\
\return\\
1.\pspace Pretend to be a user of this product and, after the string Question:, create a question based on the provided document, such as an actual problem that occurred, something you want to ask, or something you don't understand. In addition to the problem itself, it may be better to include what you are trying to do, a description of the situation, and information useful for troubleshooting (for example, the execution environment, version, execution command, error details), etc.\return\\
2.\pspace Next, assume the role of an elite contact center employee of the company that handles this product, and after the string Answer:, politely answer the beginner user's question based on the document.\return\\
3.\pspace Finally, after the string Citation:, copy and paste the exact passage from the document that forms the basis of your answer (do not make any changes; extract it as is).\return\\
\return\\
Do not create multiple questions or answers. Also, do not use any format other than Question:, Answer:, and Citation:.\return\\
\return\\
Now, please create the question in Japanese based on the information in the document below.\return\\
Document:\ptemplate{chunk}
\end{tcolorbox}
    \caption{Prompt used for synthesizing JP1-QA. Chunks extracted from JP1 manuals are inserted to \textit{\{chunk\}}. We use the Japanese prompt in our experiments.}\label{fig:synthesis_prompt_factoidQA}
\end{figure}

%% file: appendix/prompt_synthesize_factoidQA.tex
\begin{figure*}[t]
    \centering
    \fontsize{7.5pt}{7.5pt}\selectfont

    \begin{tcolorbox}
以下の「\#\#\# Document」には、ITシステムを制御・管理するためのミドルウェアであるJP1に関する説明が記載されています。\return\\
「\#\#\# Document」の文章を言い換えることで質問と回答のペアを1つ作ってください。\return\\
\return\\
ただし、必ず下記の条件を守ってください。\return\\
・回答が必ず「\#\#\# Document」の文章に記載されている1つの単語となる\return\\
・「\#\#\# 専門文書」の内容に基づいて、必ず回答可能な質問である\return\\
・回答が一意に定まる\return\\
・質問は「\#\#\# Question」の後に続けて書く\return\\
・回答は「\#\#\# Answer」の後に続けて書く\return\\
\return\\
下記は作成例です。
\return\\
\#\#\# Document\return\\
(2)\pspace計画実行登録\return\\
計画実行登録は，ジョブネットのスケジュール定義やジョブネットが属するジョブグループのカレンダー情報に基づいて実行予定をスケジュールします。
計画実行登録の場合，実行登録後は初回のジョブネットの実行予定だけが確定されたスケジュールで，それ以降のスケジュールは擬似予定（シミュレーションされたスケジュール）という扱いになります。擬似予定については，「4.4.2(1) スケジュールシミュレーション」を参照してください。次回の実行予定は，前回の実行予定のジョブネットが開始された時点でスケジュール確定します。\return\\
\return\\
\#\#\# Question\return\\
ジョブネットのスケジュール定義やジョブネットが属するジョブグループのカレンダー情報に基づいて実行予定をスケジュールする方法は何と言いますか？\return\\
\return\\
\#\#\# Answer\return\\
計画実行登録\return\\
\return\\
\return\\
それでは、下記の「\#\#\# Document」に対して上述の条件を守って、質問と回答をつくってください。絶対に質問と回答以外を出力してはいけません。解説も不要です。\return\\
\return\\
\#\#\# Document\return\\
\ptemplate{fact}\return\\
\return\\
\tcblower
\textcolor{gray}{(A prompt translated into English)}\\
The following ``\#\#\# Document'' contains an explanation of JP1, middleware for controlling and managing IT systems.\return\\
Create one question–answer pair by paraphrasing the sentences in ``\#\#\# Document.''\return\\
\return\\
Be sure to observe the following conditions.\return\\
・The answer must be a single word that appears in the sentences of ``\#\#\# Document.''\return\\
・The question must be answerable based on the content of ``\#\#\# Document.''\return\\
・The question must have only one answer.\return\\
・Write the question immediately after ``\#\#\# Question.''\return\\
・Write the answer immediately after ``\#\#\# Answer.''\return\\
\return\\
Below is an example.\return\\
\return\\
\#\#\# Document\return\\
(2)\pspace Planned execution\return\\
In planned execution, the jobnet is scheduled based on its schedule definition and the calendar information set for the job group to which the jobnet belongs.\return\\
When you register a jobnet for planned execution, only its first execution is fixed, and subsequent executions are treated as dummy runs (simulated schedules). For details on dummy runs, see 4.4.2(1) Schedule simulation. The jobnet's next run is finalized when the current run starts.\return\\
\return\\
\#\#\# Question\return\\
What is the execution method called that sets schedules for a jobnet based on the jobnet’s schedule definitions and the calendar information of the job group to which the jobnet belongs?\return\\
\return\\
\#\#\# Answer\return\\
Planned execution\return\\
\return\\
\return\\
Now, based on the above conditions, create one question–answer pair for the ``\#\#\# Document'' below. You must output nothing other than the question and answer. No explanation is needed.\return\\
\return\\
\#\#\# Document\return\\
\ptemplate{fact}\return\\
\return\\
\end{tcolorbox}
    \caption{Prompt used for synthesizing closed-book QAs used in the knowledge evaluation. The oracle facts are inserted to \textit{\{fact\}}. We use the Japanese prompt in our experiments.}\label{fig:synthesis_prompt_jp1QA}
\end{figure*}

%% file: appendix/prompt_guidance_information.tex
\begin{figure}[t]
    \centering
    \fontsize{7.5pt}{7.5pt}\selectfont

    \begin{tcolorbox}
\#\# 回答戦略について\return\\
上記の「\#\#\# 質問文」に回答するには以下の回答戦略に遵守する必要があります。\return\\
\ptemplate{conclusions}
\return\\
それでは「\#\#\# 質問文」への回答を作成してください。\return\\
\tcblower
\textcolor{gray}{(A text translated into English)}\\
\#\# About the Answer Strategy\return\\
To answer the above ``\#\#\# Question,'' you must adhere to the following answer strategy.\return\\
\ptemplate{conclusions}\return\\
\return\\
Now, please create your answer to the ``\#\#\# Question.''\return\\
\end{tcolorbox}
    \caption{Texts inserted into prompts when using oracle conclusions that specify the reasoning direction. We itemize such oracle conclusions and fill \textit{\{conclusions\}} with them. After that, we insert this filled texts into prompts. We use the Japanese text in our experiments.}\label{fig:prompt_guidance_info}
\end{figure}

%% file: appendix/prompt_llm_as_a_judge.tex
\begin{figure}[t]
    \centering
    \fontsize{7.5pt}{7.5pt}\selectfont

    \begin{tcolorbox}
「\#\# 評価対象」に書かれている文章が「\#\# 評価基準」で示す要件を満たしているかを判断してください。\return\\
ただし、書かれていない内容の推測はせず、文章内の文字列に基づき客観的に「\#\# 評価基準」と照らし合わせてください。\return\\
判断結果を「\#\# 回答」に続けて、YesかNoで回答してください。絶対にYesかNo以外で回答してはいけません。\return\\
以上の指示に厳密に正確に従ってください。\return\\
\return\\
\#\# 評価対象\return\\
\ptemplate{generated\_answer}\return\\
\return\\
\#\# 評価基準\return\\
\ptemplate{criteria}\return\\
\return\\
\#\# 回答\return\\
\tcblower
\textcolor{gray}{(A prompt translated into English)}\\
Determine whether the text written under ``\#\# Evaluation Target'' satisfies the requirements indicated under ``\#\# Evaluation Criteria.''\return\\
However, do not infer any unstated content; instead, objectively compare the strings in the text with the ``\#\# Evaluation Criteria.''\return\\
Write your judgment result immediately after ``\#\# Answer'' and respond with either Yes or No. You must not answer with anything other than Yes or No.\return\\
Follow the above instructions strictly and precisely.\return\\
\return\\
\#\# Evaluation Target\return\\
\ptemplate{generated\_answer}\return\\
\return\\
\#\# Evaluation Criteria\return\\
\ptemplate{criteria}\return\\
\return\\
\#\# Answer\return\\
\end{tcolorbox}
    \caption{Prompt used for LLM-as-a-judge. We insert a generated answer into \textit{\{generated\_answer\}} and each condition of checklists into \textit{\{criteria\}}. We use the Japanese prompt in our experiments.}\label{fig:prompt_llm_as_a_judge}
\end{figure}

%% file: appendix/prompt_eval_elicit_capability.tex
\begin{figure}[t]
    \centering
    \fontsize{7.5pt}{7.5pt}\selectfont

    \begin{tcolorbox}
統合システム運用管理向けソフトウェア・ミドルウェア製品であるJP1に関する一問一答形式の質問が「\#\#\# Question」に記載されています。\return\\
「\#\#\# Answer」に続けて、質問の答えとなる名詞を回答してください。\return\\
\return\\
ただし、絶対に質問の答えとなる名詞のみを簡潔に回答してください。それ以外は回答してはいけません。解説も不要です。\return\\
下記は回答例です。\return\\
\return\\
\#\#\# Question\return\\
JP1/IM - ManagerがJP1/Base（イベントサービス）からJP1イベントを取得する際の条件を設定するフィルターを何といいますか？\return\\
\return\\
\#\#\# Answer\return\\
イベント取得フィルター\return\\
\return\\
\#\#\# Question\return\\
JP1製品のうち、複数の業務の内容と実行順序を定義することで定型的・定期的な業務を自動化することを目的とした製品は何ですか？\return\\
\return\\
\#\#\# Answer\return\\
JP1/Automatic Job Management System 3\return\\
\return\\
それでは、下記の「\#\#\# Question」に回答してください。\return\\
\return\\
\#\#\# Question\return\\
\ptemplate{question}\return\\
\tcblower
\textcolor{gray}{(A prompt translated into English)}\\
A question in a Q\&A format about JP1, an integrated system operations management middleware product, is provided under ``\#\#\# Question.''\return\\
Please write the noun that answers the question immediately after ``\#\#\# Answer.''\return\\
\return\\
Be sure to answer concisely with only the noun that serves as the answer to the question. Do not output anything else. No explanation is needed.\return\\
Below is an example.\return\\
\return\\
\#\#\# Question\return\\
What is the filter called that specifies the conditions under which JP1/IM – Manager acquires JP1 events from JP1/Base (event service)?\return\\
\return\\
\#\#\# Answer\return\\
Event acquisition filter\return\\
\return\\
\#\#\# Question\return\\
Among JP1 products, which product is designed to automate routine and periodic tasks by defining the contents and execution order of multiple tasks?\return\\
\return\\
\#\#\# Answer\return\\
JP1/Automatic Job Management System 3\return\\
\return\\
Now, please answer the ``\#\#\# Question'' below.\return\\
\return\\
\#\#\# Question\return\\
\ptemplate{question}\return\\
\end{tcolorbox}
    \caption{Prompt used for evaluating the elicitation capability in our knowledge evaluation. We insert synthesized a closed-book QA into \textit{\{question\}}. We use the Japanese prompt in our experiments.}\label{fig:prompt_eval_elicit_capability}
\end{figure}